\documentclass[twoside]{article}

\usepackage{caption}
\usepackage{graphicx}

\usepackage{amsthm}
\usepackage{amsmath}
\usepackage{amsfonts}
\usepackage{amssymb}
\usepackage[ruled,vlined, algo2e]{algorithm2e}

\newtheorem{assumption}{Assumption}
\newtheorem{theorem}{Theorem}
\newtheorem{corollary}{Corollary}
\newtheorem{lemma}{Lemma}

\usepackage[accepted]{aistats2022}
\begin{document}



\twocolumn[

\aistatstitle{Preconditioned Federated Learning}

\aistatsauthor{Zeyi Tao \And Jindi Wu \And Qun Li}

\aistatsaddress{\\ Department of Computer Science\\ William \& Mary\\ Williamsburg, VA, USA\\ ztao, jwu21, liqun@wm.edu}]

\begin{abstract}
Federated Learning (FL) is a distributed machine learning approach that enables model training in communication efficient and privacy-preserving manner. The standard optimization method in FL is Federated Averaging (FedAvg), which performs multiple local SGD steps between communication rounds. FedAvg has been considered to lack algorithm adaptivity compared to modern first-order adaptive optimizations. In this paper, we propose new communication-efficient FL algortithms based on two adaptive frameworks: local adaptivity (PreFed) and server-side adaptivity (PreFedOp). Proposed methods adopt adaptivity by using a novel covariance matrix preconditioner. Theoretically, we provide convergence guarantees for our algorithms. The empirical experiments show our methods achieve state-of-the-art performances on both i.i.d. and non-i.i.d. settings.
\end{abstract}

\section{Introduction}
Federated learning (FL)~\cite{mcmahan2017communication} is a distributed learning paradigm that collaboratively trains models in heterogeneous networks without data sharing.
FL performs the model update in parallel on a small subset of the total clients and averages the local models on a centralized server only once in a while.
% In FL, the training data is never shared with a centralized server or other clients. 
% Hence, FL is more appealing than traditional centralized learning due to security and privacy reasons. 

FL has two fundamental characters that distinguish it from the standard parallel optimization. 
First, the training data is non-iid~\cite{li2019convergence,li2020federated}. 
In FL, the data is massively distributed over different clients such as banks, hospitals, and information institutions~\cite{kairouz2019advances}. 
Data shard on each client is sampled from a possibly different distribution (a.k.a., heterogeneous or non-i.i.d. data samples). Consequently, local data fail to represent the overall distribution. Data heterogeneity does not only bring challenges to algorithm design but also makes theoretical analysis much harder. 
Second, considerable clients are involved in FL training ranging from smartphones, network sensors to powerful servers. 
The standard distributed optimizations, such as stochastic gradient descent (SGD), may not be suitable for FL due to the high communication cost of iterative gradient methods. 

In order to handle heterogeneity and tackle high communication costs, 
communication-efficient optimization methods are widely adopted. 
Federated averaging (FedAvg)~\cite{mcmahan2017communication} is the first and perhaps the most widely used FL algorithm. 
FedAvg runs $K$ steps of SGD in parallel on a small sampled subset of devices before performing global model synchronization. 
In this way, FedAvg can significantly reduce the amount of communication required to train a model. 

While FedAvg has demonstrated empirical success in certain applications, recent works have highlighted its slow convergence issues in heterogeneous settings. 
This is due to (1) model divergence~\cite{li2018federated,karimireddy2019scaffold}, where locally trained models drift away from globally optimal models causing slow and unstable convergence, and (2) a lack of adaptivity~\cite{reddi2020adaptive}. 
FedAvg updates local models via vanilla SGD, resulting in poor performance if we train models with heavy-tail stochastic gradient noise distributions such as language models.
~\cite{zhang2019adam} theoretically have showed that the preconditioned methods such as AdaGrad~\cite{duchi2011adaptive}, Adam~\cite{kingma2014adam}, and AMSGrad~\cite{reddi2019convergence} have better performance in such settings. 
Spurred by the advances of preconditioned methods,~\cite{reddi2020adaptive} proposed the first decentralized Adam method (FedAdam) directly applied to federated learning.

In this work, we propose preconditioned federated algorithms based on two different adaptive frameworks: client-side~\cite{xie2019local} and server-side~\cite{reddi2020adaptive}. 
We first propose Preconditioned Federated (PreFed) algorithm, a client-side adaptive federated optimization that accelerates the client models training. 
In addition, we propose PreFedOpt algorithm, where the adaptivity is applied only on the server. 
For both proposed algorithms, we achieve adaptivity by using a covariance matrix-based preconditioner~\cite{ida2016adaptive}. 
We periodically synchronize this preconditioner over all clients on a centralized server.

A key insight we have in developing proposed algorithms is that covariance matrix preconditioner reduces the noise in the stochastic gradient-based approximation of Fisher information~\cite{amari1998natural}. Therefore the curvature in the loss landscape can be precisely captured by noise-reduced Fisher information to accelerate the gradient descent training~\cite{dauphin2015equilibrated,ida2016adaptive,balles2018dissecting}.

In light of the above, we highlight the main contributions of the paper. 
We propose preconditioned FL algorithms (PreFed and PreFedOpt) based on two adaptive federated optimization frameworks.
We provide convergence guarantees for our methods for non-convex online FL optimizations. 
Despite encouraging theoretical results, our empirical evaluation indicates that our methods consistently outperform recent state-of-the-art adaptive federated algorithms.

% \begin{algorithm}[t]
% \caption{P-FEDADAM}\label{algo1}
% \begin{algorithmic}[1]
% \STATE Initialization: $x_0$, learning rate $\eta_0=1e^{-3}$, decay parameter $\beta_1, \beta_2 \in (0,1)$
% \STATE \textbf{for} t = 0 \textbf{to} T-1 \textbf{do}
% \STATE \quad Sample subeset $\mathcal S$ of clients
% \STATE \quad $x_{i, 0}^t = x_t$
% \STATE \quad $v_{i,0}^t = \frac{1}{|\mathcal S|} \sum_{i \in \mathcal S} v_{i}^{t-1}$
% \STATE \quad\quad \textbf{for} each client $i\in \mathcal S$ \textbf{in parallel do}
% \STATE \quad\quad\quad \textbf{for} k = 0 \textbf{to} K-1 \textbf{do}
% \STATE \quad\quad\quad\quad $g_{i,k}^t = \nabla F_i(x_{i,k}^t; d_i)$
% \STATE \quad\quad\quad\quad $m_{i,k}^t = \beta_1 m_{i,k-1}^t + (1-\beta_1)g_{i,k}^t$
% \STATE \quad\quad\quad\quad $v_{i,k}^t = \beta_2 v_{i,k-1}^t + (1-\beta_2)g_{i,k}^t\odot g_{i,k}^t$
% \STATE \quad\quad\quad\quad $x_{i, k+1}^t = x_{i, k}^t - \frac{\eta_t}{\sqrt{v_{i,k}^t}+ \epsilon} m_{i,k}^t$
% \end{algorithmic}
% \end{algorithm}

% \begin{algorithm}[t]
% \caption{COV-P-FEDADAM}\label{algo2}
% \begin{algorithmic}[1]
% \STATE Initialization: $x_0$, learning rate $\eta_0=1e^{-3}$, decay parameter $\beta_1, \beta_2 \in (0,1)$
% \STATE \textbf{for} t = 0 \textbf{to} T-1 \textbf{do}
% \STATE \quad Sample subeset $\mathcal S$ of clients
% \STATE \quad $x_{i, 0}^t = x_t$
% \end{algorithmic}
% \end{algorithm}
\section{Related Work}
There is an enormous amount of work studying standard federated optimization\footnote{By saying standard federated optimization, we refer to the non-adaptive federated optimization such as FedAvg~\cite{mcmahan2017communication}, SCAFFOLD~\cite{karimireddy2019scaffold}, etc.}~\cite{li2018federated,li2019convergence,haddadpour2019convergence} and first-order preconditioning~\cite{dauphin2015equilibrated,qu2020diagonal,bischoff2021second,li2018preconditioner}, which is too much to discuss here in detail. 
Thus, we only briefly discuss two lines of work that are most relevant to our paper here. 
First, adaptive federated optimzation e.g.,~\cite{xie2019local,reddi2020adaptive}, to understand and give theoretical guarantees. 
Second, the covariance matrix-based gradient descent preconditioning~\cite{ida2016adaptive}.

\textbf{Adaptive Federated Learning} In order to solve the federated learning problem~\cite{mcmahan2017communication} first proposed FedAvg as a communication efficient FL algorithm. 
FedAvg runs vanilla SGD on client model updates and averages the client models to produce the new global model for next round training.
\begin{algorithm2e}[t]
\caption{FedAdaGrad~\cite{reddi2020adaptive}}
\label{alg:fedadagrad}
\SetKwInOut{KwIn}{Clients (ClientOpt)}
\SetKwInOut{KwOut}{Server (ServerOpt)}
Initialization: $w_0, v_0, \eta_0, \epsilon>0$\;
\For{$t=0, 1, \cdots, T-1$}{
\KwIn{}
Sample subset $\mathcal S$ of clients\;
$w_{i, 0}^t = w_t$\;
\For{each client $i\in \mathcal S$ in parallel}{
\For{$k=0, \cdots, E-1$}{
$g_{i,k}^{t} = \nabla F_i(w_{i,k}^{t}, x_i), x_i \sim \mathcal D_i$\;
$w_{i,k}^{t} = w_{i,k-1}^{t} - \eta_{t} g_{i,k}^{t}$\;
}
$\Delta_i^t = w_{i,K}^{t} - w_t$
}
\KwOut{}
$\Delta_{t} = \frac{1}{|\mathcal S|} \sum_{i\in \mathcal S} \Delta_i^t$\;
$v_t = v_{t-1} + \Delta_{t}^2$\;
$w_{t+1} = w_{t} - \frac{\eta_t}{\sqrt{v_t} + \epsilon} \odot \Delta_{t}$\;
Server broadcast $w_{t+1}$ to all clients\;
}
\end{algorithm2e}
To accommodate the first-order adaptive methods, ~\cite{xie2019local} proposed AdaAlter, a method for FL that allows the client models to use adaptive learning rates. 
AdaAlter uses the same adaptive term as AdaGrad\footnote{Adagrad uses the square root of the sum of the outer product of the past gradients to achieve adaptivity such as $\sum_{i=1}^{t}g_ig_i^{\top}$ at $t$-th round} or its variants and simultaneously performs lazy adaptive term updates on global model synchronization. 
AdaAlter has been proven that achieves $\mathcal O(1/\sqrt{T})$ convergence rate on federated optimization. 
However, the empirical result on the 1 billion words dataset shows that AdaAlter highly depends on the adaptive term synchronization, which makes communication inefficient.

Recently,~\cite{reddi2020adaptive} developed the first generic framework for federated optimization using adaptive first-order gradient descent algorithms named FedOpt. 
FedOpt has two optimizers ClientOpt and ServerOpt. 
ClientOpt, similar to FedAvg, performs parallel SGD on the client model.
While the SeverOpt adopts adaptive optimization (one of AdaGrad, YOGI, or Adam) for parameter updates on the server.
Further, is has been shown that the AdaAlter is a special case of FedOpt. 
% The critical difference between FedOpt and AdaAlter is that FedOpt performs adaptive steps only on the server. 
FedOpt, compared to the AdaAlter, avoids aggregating optimizer states across clients, making it more communication efficient. 
We illustrate FedAdaGrad in Algorithm~\ref{alg:fedadagrad} as a reference. 

\textbf{Covariance matrix-based SGD preconditioning} 
Recently, many works have focused on preconditioned SGD methods~\cite{dauphin2015equilibrated,ida2016adaptive,staib2019escaping,qu2020diagonal}. 
The preconditioners can be categorized into the Newton type and the Fisher type~\cite{amari1998natural}. 
The Newton type is closely related to the Newton method, and is suitable for small and middle scale machine learning problems due to computational efficiency reasons. 
The Fisher type preconditioner that uses the inverse of Fisher information matrix is more appealing because the Fisher information metric can be well defined in most learning problems. 

The empirical success of adaptive methods such as Adam could be explained by using preconditioned SGD.
In particular, the adaptivity of Adam is an approximation to the Fisher information matrix by accumulating first-order gradients in the past.
To acquire a better approximation of Fisher information, ~\cite{ida2016adaptive} proposed covariance matrix-based SGD preconditioning (SDProp) by reducing the noise of gradient covariance.
Empirical results of SDProp have shown that effectively handling the noise present in the gradients can improve the network's performance on image classification and language tasks. 
Nerveless, SDProp lacks theoretical analysis on convergence.

In this work, we apply the covariance matrix-based preconditioner to FL to accelerate the training and improve the overall model performance.
To our best knowledge, there is no preconditioning method has been applied in FL optimization so far. 
Our method is perhaps the first one. 
% In this paper, our proposed method follows the same design pattern as AdaAlter but is different in adaptive term design. 
% We adopt the covariance matrix-based preconditioning in ~\cite{ida2016adaptive} to perform adaptivity. 
% In our design, we compute variance reduced Fisher information across all clients to accelerate the training. 
% To our best knowledge, it is worth mentioning that no preconditioning methods have been applied in FL optimization so far. 
% Our method is perhaps the first one. 

\section{Preliminaries}
\subsection{Notations and Problem Formulation}
For $x, y \in R^d$, let $\sqrt{x}, x\odot y$, and $x/y$ denote the element-wise square root, multiplication, and division of the vectors. For $w\in R^d$, we use both $w_{i,j}$ and $[w_i]_j$ to denote its $j$-th coordinate for $i$-th client model.

In FL, we assume that there are totally $m$ clients which hold their private data and
perform local computation, as well as a central server which sends and receives messages from the clients. FL aims to minimize the following functions:
%FL typically aims to minimize the empirical risk over heterogeneous data distributed across multiple clients:
\begin{equation}\label{fed}
\min_{w} F(w) = \frac{1}{m}\sum_{i=1}^{} F_i(w)
\end{equation}
where $F_i(w) = \mathbb E_{z\sim \mathcal D_i} [f_i(w, z)]$, is the local objective function of the $i$-th client and $\mathcal D_i$ is the data distribution for $i$-th client. For $i\neq j$, $\mathcal D_i$ and $\mathcal D_j$ may be very different. Suppose the $i$-th client has $n_i = |\mathcal D_i|$ training data: $x_{i, 1}, x_{i, 2}, \cdots x_{i, n_i}$. The local objective $F_i(\cdot)$ is defined by
\begin{equation}
F_i(w) \triangleq \frac{1}{n_i} \sum_{j=1}^{n_i} f_i(w; x_{i, j})
\end{equation}
where $f(\cdot; \cdot)$ is a loss function that could be convex or nonconvex.

To solve the federated learning problem (Equation~\ref{fed}), FedAvg~\cite{mcmahan2017communication} is used. In FedAvg, the central server broadcasts the latest model (say the $t$-th round) $w_t$ to all the clients. Then every client (say the $i$-th) let $w_i^t = w_t$ and performs $K (K\geq0)$ local updates:
\begin{equation}\label{fed-local-sgd}
    w_{i, k}^{t} = w_{i, k-1}^{t} - \eta_{t} g_{i, k}^{t}
\end{equation}
where $k = 0, 1, \cdots, K-1$ is local iteration, $\eta_{t}$ is the learning rate (a.k.a. step size).
$g_{i,k}^{t} = \nabla F_i(w_{i,k}^{t}, x_{i})$ where $x_i$ is a sample uniformly chosen from the local data $\mathcal D_i$. After $K$ local updates, the central server randomly sampled subset of client models $\mathcal S$, $w_{1,K}^{t},\cdots, w_{s,K}^{t}$ and a aggregates them to produce the new global $w_{t+1}$. 

\subsection{Adaptive Federated Optimization}
\textbf{Client-side Adaptivity} Adding adaptivity to the above settings of FL is straightforward. In~\cite{xie2019local} work (AdaAlter), instead of using vanilla SGD on the local client like Equation~\ref{fed-local-sgd}, AdaAlter performs adaptive local updates:
\begin{equation}
    \text{(on Client i)}\quad w_{i,k}^{t} = w_{i,k-1}^{t} - \frac{\eta}{\sqrt{v_{i,k}^{t}} + \tau} \odot g_{i,k}^{t}
\end{equation}
where $\tau$ controls the algorithms’ degree of adaptivity~\cite{reddi2020adaptive}.
$v_{i,k}^{t}$ is AdaGrad-liked preconditioner defined as $v_{i,k}^{t} = v_{i,k-1}^{t} + g_{i,k}^{t} \odot g_{i,k}^{t}$. AdaAlter synchronizes model parameters periodically as standard FedAvg does. In addition, AdaAlter updates preconditioner such that $v_{t+1} = \frac{1}{s} \sum_{i\in \mathcal S} v_{i,K}^{t}$ over all participating clients.

\textbf{Server-side Adaptivity}~\cite{reddi2020adaptive} proposed FedOpt (or FedAdaGrad) that applies the adaptivity when the central server performs global model fusion e.g.,
\begin{equation}
    \text{(on Server)}\quad w_{t+1} = w_t - \frac{\eta}{\sqrt{v_t} + \tau}\odot \Delta_t
\end{equation}
where $\Delta_t = \sum_{i=1}^{m}w^{t}_{i,K} - w_t$ and $v_t = v_{t-1} + \Delta_t\odot \Delta_t$.
Algorithm~\ref{alg:fedadagrad} provides more details about FedAdaGrad.
Clearly, FedAdaGrad does not require extra communication, and one can directly apply FedAdaGrad to the current FL without any further modification. 

\subsection{Preconditioned SGD}
Adaptive learning rate algorithms are widely used for the efficient training of deep neural networks. 
The empirical success of adaptive methods could be explained by using preconditioned SGD~\cite{dauphin2015equilibrated}. 
In precondition, we seek to locally transform the curvature of the loss function so that its curvature is equal in all directions. 
A direct benefit of this is gradient descent method could escape from the saddle point much easier, which has been recently found as a major obstacle in machine learning training~\cite{choromanska2015loss}.

Preconditioned SGD works in this way.
Consider a loss function $f$ with parameter $w \in R^d$.
We introduce a linear transform of variables $\hat{w} = P^{\frac{1}{2}}w$ with a non-singular matrix $P^{\frac{1}{2}}$.
Now, instead of performing optimization on $f(w)$, we focus on variable transformed function $\hat{f}(\hat{w})$.
The gradient and hessian of $\hat{f}(\hat{w})$ with respect to $w$ are $P^{-\frac{1}{2}} \nabla f(w)$ and $P^{-\frac{1}{2}} H P^{-\frac{1}{2}}$. 
Here $H = \nabla^2 f(w)$ is Hessian matrix.
Consequently, the gradient descent for optimizing $\hat{f}(\hat{w})$ transform to:
\begin{equation}\label{conditionupdate}
w_{t} = w_{t-1} - \eta_t P_t^{-1} \nabla f(w)
\end{equation}
with respect to the original variable $w$. Particularly, when $P^{\frac{1}{2}} = H^{\frac{1}{2}}$, the new Hessian $\nabla^2 \hat{f}(\hat{w}) = I$ is perfectly conditioned and Equation~\ref{conditionupdate} corresponds the Newton method.
Searching for perfect conditioner $P^{\frac{1}{2}} = H^{\frac{1}{2}}$ is prohibited in modern neural networks because $H^{\frac{1}{2}}$ exists only when $H$ is positive-semidefinite and has high computational complexity (e.g.$\mathcal O(d^3)$). 

To avoid using Hessian to obtain the curvature information, adaptive methods like AdaGrad and RMSProp use the Fisher-type preconditioner $P_t = \sqrt{\sum_{i=1}^{t} g_ig_i^{\top}}$ and $P_t = \sqrt{(1-\beta)\sum_{i=1}^{t} \beta^{t-i}g_ig_i^{\top}}$ respectively.\footnote{In RMSProp $\beta > 0$ is chosen appropriately such that $\sqrt{(1-\beta)\sum_{i=1}^{t} \beta^{t-i}g_ig_i^{\top}} \approx \mathbb E[g_tg_t^{\top}]$} 
The Fisher information can also capture curvature information in the loss landscape if the objective function is a log-likelihood~\cite{staib2019escaping}.
In most cases, this condition holds where we often use cross-entropy as our loss function.
Hence, $\mathbb E[gg^{\top}]$ approximates the Fisher information matrix, which present curvature information in the parameter space.

However, in stochastic optimization approaches such as mini-batch settings, the first-order gradients always contain noise.
Consequently, the square root of the uncentered variances in $P_t$ contains noise, producing an approximation of the Fisher information matrix with noise.

\section{Proposed Method}
In this section, we discuss the motivation behind our proposed methods. 
The key insight of our proposed method is that we construct more precise Fisher information by using a noise-reduced covariance matrix. 
Then we present two new adaptive federated algorithms based on two adaptive FL optimization frameworks. 

\subsection{Motivation}
In general, consider an objective function $F(w)$ with parameter $w$. The preconditioned SGD has the following updated rule:
\begin{equation}
\small
\begin{split}
    w_{t} &= w_{t-1} - \eta P_t^{-1} g_t\\
    &= w_{t-1} - \eta P_t^{-1} \nabla F(w_t) - \eta P_t^{-1} (g_t -\nabla F(w_t))
\end{split}
\end{equation}
where $\nabla F(w)$ is true gradient at $w$ and $g_t= \nabla F(w_t, d)$ is noised gradient.
The $P_t^{-1} \nabla F(w_t)$ term is deterministic. 
The $P_t^{-1} (g_t -\nabla F(w_t))$ is measuring the stochastic gradient noise.
By letting stochastic gradient $g_t$ with mean of $\mathbb E[g_t] =\nabla F(w)$ and assume $P_t = \mathbb E[g_tg_t^{\top}]^{1/2}$ where $\mathbb E[g_tg_t^{\top}]$ is non-singular.
Based on the recent study~\cite{staib2019escaping}, the stochastic noise $\zeta_t = P_t^{-1} (g_t -\nabla F(w_t))$ has mean $\mathbb E[\zeta_t] = 0$ and covariance:
\begin{equation}\label{cov}
    \text{Cov}(\zeta_t) = I - P_t^{-1} \nabla F(w_t) \nabla F(w_t)^{\top} (P_t^{-1})^{\top}
\end{equation}
Given the adaptive setting above, the covariance $\text{Cov}(\zeta_t)$ is approximately the identity matrix $I$ for all $t \in [T]$.
% When the model approaches the stationary point, $\nabla F(w_t)$ is close to zero.
% The covariance $\text{Cov}(\zeta_t)$ is s approximately the identity matrix $I$.

This observation demonstrates one of the reasons for the success of precondition (adaptive) methods on non-convex optimization.
It is well-known that gradient descent methods make slow progress when the curvature of the loss function is very different in separate directions~\cite{dauphin2015equilibrated}.
When Equation~\ref{cov} is an identical matrix, it ensures the gradient noise is approximately isotropic. The curvature of the optimization surface is equal in all directions making the optimizer quickly escapes from the saddle point.

Intuitively, the noise of stochastic gradient is even higher due to heterogeneous settings. 
Such high gradient variance may cause non-isotropic curvature slowing down the training. 
\textit{Can we reduce gradient variance across all client gradients via precondition?} 
This is the primary motivation behind our approach. 
In addition, our approach should be computation and communication efficient. 
In the next section, we present our preconditioned FL optimization algorithms.

\subsection{Preconditioned Federated Optimization}
First, it is natural to assume the first-order gradients follow Gaussian distributions in FL due to the recent observations on probabilistic modeling~\cite{sra2012optimization,andrychowicz2016learning,zhang2018noisy}. By following~\cite{ida2016adaptive}, we assume the gradient on $i$-th client $g_{i, t} = \nabla F_i(w_t, d_i) \in R^d$:
\begin{equation}
    g_{i,t} \sim \mathcal N(\bar{g}_t, 	C_{i,t}) \quad \forall i \in [m]
\end{equation}
where $\bar{g}_t = \nabla F(w_t)$ is the true gradient without the noise. $\mathcal N(\bar{g}_t, C_{i,t})$ is a Gaussian distribution with mean $\bar{g}_t$ and covariance matrix $C_{i,t}$ and $C_{i,t}$ is the covariance matrix of $g_{i,t}$ whose size is $d \times d$. 
Suppose we use $m$ clients to perform synchronization.
Then we have $m$ normally distributed gradients such as $g_{1,t}, \cdots, g_{m,t}$.
The distribution of averaged gradient over $m$ clients of $G_t = \frac{1}{s}\sum_{i\in\mathcal S} g_{i,t}$ is given
\begin{equation}
    G_t \sim \mathcal N(\bar{g}_t, \frac{1}{m^2}\Sigma_{t})
\end{equation}
where $\Sigma_{t} = C_{1,t} + \cdots + C_{m,t}$ is sum of $m$ covariance matrices.
$\Sigma_{t}$ has the same size of covariance matrices as $d \times d$.
Since $C_{m, t}$ is assumed to be a positive semi-definite matrix, $\Sigma_{t}$ is also positive semi-definite matrix.
Later we can decompose $\Sigma_t$ as $\Sigma_t = VDV^{\top}$ where $V$ is an orthogonal matrix of $d \times d$ and $D$ is a diagonal matrix of eigenvalues.

We define preconditioned gradient $\hat{G}_t = P_t^{-1} G_t$. The distribution of transformed gradient changes to
\begin{equation}\label{pretrans}
    \hat{G}_t \sim \mathcal N(P_t^{-1}\bar{g}_t, \frac{1}{m^2}P_t^{-1}\Sigma_{t}(P_t^{-1})^{\top})
\end{equation}
Equation~\ref{pretrans} is hold due to the linear transform of random variable from normal distribution. 
If random variable $X\sim \mathcal N(\mu, \Sigma)$ and $Y = AX + B$, then $Y\sim \mathcal N(A\mu, A\Sigma A^{\top} + \Sigma_B)$.
Now, we can reduce the covariance of $\hat{G}_t$ to $\frac{1}{m^2} I$ by letting $P_t = \Sigma_t^{\frac{1}{2}}$ such that
\begin{equation}\label{covreduced}
\small
    \begin{split}
        \hat{G}_t &\sim \mathcal N(P_t^{-1}\bar{g}_t, \frac{1}{m^2}P_t^{-1}\Sigma_{t}(P_t^{-1})^{\top})\\
         &\sim \mathcal N(\Sigma^{-\frac{1}{2}}\bar{g}_t, \frac{1}{m^2}\Sigma^{-\frac{1}{2}}\Sigma_{t}(\Sigma^{-\frac{1}{2}})^{\top})\\
         &\sim \mathcal N(\Sigma^{-\frac{1}{2}}\bar{g}_t, \frac{1}{m^2}VD^{-\frac{1}{2}}V^{-1}VDV^{-1}(VDV^{-1})^{\top})\\
         &\sim \mathcal N(\Sigma^{-\frac{1}{2}}\bar{g}_t, \frac{1}{m^2} I)
    \end{split}
\end{equation}
Third line is due to orthogonal matrix $V$ with $VV^{\top} = I$ and $(VDV^{-1})^{\top} = VDV^{-1}$.
Equation~\ref{covreduced} indicates that the  transformation of $\hat{G}_t = \Sigma_t^{-\frac{1}{2}} G_t$ result in the Gaussian distribution of $\hat{G}_t$ whose covariance matrix is scaled identity matrix $\frac{1}{m^2}I$. 
When we increase the number of participated clients, the gradient variance decreases as $\mathcal O (1/m^2)$. 

In this way, we reduce the preconditioned gradient noises and ensure they are equal in all directions.
We introduce the basic version online updating rule of preconditioned federated optimization, and we leave more discussion in the next section.
We compute the first order gradients parallelly at each local iteration on each client as $g_{i,k}=\nabla F_i(w_{i,k}^t)$. We compute averaged gradient
\begin{equation}
m_{k}^{t} = \frac{1}{|\mathcal S|}\sum_{i\in\mathcal S}g_{i,k}
\end{equation}
over all selected clients.
The local covariance based preconditioner is computed according to
\begin{equation}\label{basicprecon}
    P_{i, k} = P_{i, k-1} + (g_{i, k} - m_{k}^{t})\odot (g_{i, k} - m_{k}^{t})
\end{equation}
We update the local model on $i$-th client at $k$-th step:
\begin{equation}\label{basicupdate}
    w_{i,k+1}^t = w_{i, k}^t - \frac{\eta_t }{\sqrt{P_{i,k}} + \tau} \odot g_{i, k} 
\end{equation}
Clearly, the update rule in Equation~\ref{basicupdate} is not communication efficient due to the synchronization in every local iteration. We address the communication issue by using (1) \textbf{lazy update of preconditioner} and (2) \textbf{server-side preconditioning}. The next section formally presents our efficient preconditioned federated optimization algorithm under the above two settings.

\subsection{PreFed and PreFedOpt}
To reduce the communication cost on preconditioned FL optimization (Equation~\ref{basicupdate}), we adopt (1) a lazy preconditioner update PreFed (Algorithm~\ref{alg:prefed}). We simultaneously update covariance preconditioner $P_t$ on global model synchronization; (2) the server-side preconditioning PreFedOpt (Algorithm~\ref{alg:prefedopt}). That is, instead of performing local preconditioning on each client model, we only apply preconditioner on global model update.

PreFed has the same design pattern as AdaAlter, in which we perform preconditioning (adaptivity) on each client and synchronize the preconditioner (adaptive term) on a server over time. Since the accumulator $m_t$ and $P_t$  are the same sizes as the model’s trainable weights, client-to-server communication doubles for PreFed, relative to FedAvg. On the other hand, the local precondition accelerates the convergence speed of the client model. The trade-off between convergence rate and communication cost is studied in section X. One simple way to reduce the communication cost is using the diagonal preconditioner. For example, we do not share the full matrices. Instead, we use diagonal covariance matrices:
\begin{equation}
    P_t = P_{t-1} + \text{diag} ((g_t - m_t)\odot (g_t - m_t))
\end{equation}
We omit the client id $i$ and local iteration $k$ for simply demonstrating the percondistioner update rule (also applies on the following Equation~\ref{emapreprefed} and~\ref{emapreprefedopt}).
Since this approach only needs the diagonal terms, the memory and computation costs are $\mathcal O(d)$. 

By contrast, PreFedOp does not require extra communication or client memory relative to PreFed. Thus, we see that server-side adaptive optimization benefits from lower per-round communication and client memory requirements, which is essential for FL applications.
In PreFed and PreFedOpt, gradient coordinates are scaled with the accumulated information of all the past squared gradients like AdaGrad. 
A significant drawback of using the AdaGrad liked preconditioner in Equation~\ref{basicprecon} is the progressive increase of its elements, which leads to a rapid decrease of the learning rate. 
To prevent this monotonically decreasing learning rate, we modify the PreFed's (also PreFedOpt) preconditioner by using exponential moving average (EMA). 
By adequately choosing decay parameter $\beta > 0$, we change preconditioner update rule~\ref{basicprecon} to:
\begin{equation}\label{emapreprefed}
    P_{t} = \beta P_{t} + (1-\beta)(g_{t} - m_{t})\odot (g_{t} - m_{t})
\end{equation}
For PreFedOpt, similarly, we perform server-side preconditioner update:
\begin{equation}\label{emapreprefedopt}
    P_t = \beta P_{t-1} + (1 - \beta)(\Delta_{t} - m_t)\odot (\Delta_{t} - m_t)
\end{equation}
Taking all these into account, we formally present our methods in Algorithm~\ref{alg:prefed} and Algorithm~\ref{alg:prefedopt}.

\begin{algorithm2e}[t]
\caption{Local Preconditioned Federated Optimization (PreFed)}
\label{alg:prefed}
\SetKwInOut{KwIn}{Clients}
\SetKwInOut{KwOut}{Server}
Initialization: $w_0, m_0, P_0, \beta_1=0.9,\beta_2=0.9,\tau>0$\;
\For{$t=0, 1, \cdots, T-1$}{
\KwIn{}
Sample subset $\mathcal S$ of clients\;
$w_{i, 0}^t = w_t, m_{i,0}^t = m_{t}, P_{i,0}^t = P_t$\;
\For{each client $i\in \mathcal S$ in parallel}{
\For{$k=0, \cdots, K-1$}{
$g_{i,k}^{t} = \nabla F_i(w_{i,k}^{t}, x_i), x_i \sim \mathcal D_i$\;
$m_{i,k}^{t} = \beta_1 m_{i,k-1}^{t} + (1-\beta_1)g_{i,k}^{t}$\;
$v_{i,k} = (g_{i,k}^{t}- m_{i,k}^{t})\odot(g_{i,k}^{t} - m_{i,k}^{t})$\;
$P_{i,k}^{t} = \beta_2 P_{i,k-1}^{t} +(1 -\beta_2)v_{i,k}$\;
$w_{i,k}^{t} = w_{i,k-1}^{t} - \frac{\eta}{\sqrt{P_{i,k}^{t}}+\tau} \odot m_{i,k}^{t}$\;
}
}
\KwOut{}
$w_{t+1} = \frac{1}{|\mathcal S|} \sum_{i\in \mathcal S} w_{i,K}^{t}$\;
$P_{t+1} = \frac{1}{|\mathcal S|} \sum_{i\in \mathcal S} P_{i,K}^{t}$\;
Server broadcast $w_{t+1}, P_{t+1}$ to all clients\;
}
\end{algorithm2e}

\begin{algorithm2e}[t]
\caption{Preconditioned Federated Optimization (PreFedOpt)}
\label{alg:prefedopt}
\SetKwInOut{KwIn}{Clients (ClientOpt)}
\SetKwInOut{KwOut}{Server (ServerOpt)}
Initialization: $w_0, m_0, P_0, \beta_1=0.9,\beta_2=0.99,\eta=0.002, \tau>0$\;
\For{$t=0, 1, \cdots, T-1$}{
\KwIn{}
Sample subset $\mathcal S$ of clients\;
$w_{i, 0}^t = w_t$\;
\For{each client $i\in \mathcal S$ in parallel}{
\For{$k=0, \cdots, E-1$}{
$g_{i,k}^{t} = \nabla F_i(w_{i,k}^{t}, x_i), x_i \sim \mathcal D_i$\;
$w_{i,k}^{t} = w_{i,k-1}^{t} - \eta_{t} g_{i,k}^{t}$\;
}
$\Delta_i^t = w_{i,K}^{t} - w_t$
}
\KwOut{}
$\Delta_{t} = \frac{1}{K|\mathcal S|} \sum_{i\in \mathcal S} \Delta_i^t$\;
$m_t = \beta_1 m_{t-1} + (\beta_1 -1)(\Delta_{t} - m_{t-1})$\;
$P_t = \beta_2 P_{t-1} + (1-\beta_2)(\Delta_{t} - m_t)\odot (\Delta_{t} - m_t)$\;
$w_{t+1} = w_{t} - \frac{\eta}{\sqrt{P_t} + \tau} \odot \Delta_{t}$\;
Server broadcast $w_{t+1}$ to all clients\;
}
\end{algorithm2e}
\section{Convergence Analysis}
We give the theoretical analysis of PreFed and PreFedOpt in this section.
In the standard federated learning Equation~\ref{fed}, we consider more realistic case that the function $F_i$ (and therefore $F$) may be non-convex. For each $i$ and $w$, we assume $g_i(w)$ is an unbiased stochastic gradient estimation of the client true gradient $\nabla F_i(x)$. In addition, we make the following assumptions.
\begin{assumption}\label{as1}
 (Bounded gradient). The function $F_i$ has bounded gradients, i.e., for any $i\in [m], x\in R^d$ we have
\begin{equation*}
    ||\nabla F_i(x) ||_j \leq G \text{\quad for all\quad} j\in[d]
\end{equation*}
\end{assumption}
\begin{assumption}\label{as2}
(L-Lipschitz smooth). The function $F_i(\cdot)$ is L-Lipschitz smooth for all $i \in [m]$, i.e.
\begin{equation*}
    ||\nabla F_i(x) - \nabla F_i(y) || \leq L||x -y ||\text{\quad for all\quad} x,y\in R^d
\end{equation*}
\end{assumption}
\begin{assumption}\label{as3}
 (Bounded variance). The function $F_i$ has $\sigma_{local}^2$-bounded local variance i.e., $\mathbb E || g_{i,j} - [\nabla F_i(w_i)]_j || = (\sigma_{local})_j^2$ for all $w_i\in R^d, j\in [d]$ and $i \in [m]$. We also assume $F_i$ has $\sigma_{global}^2$-bounded global variance i.e., $\mathbb E|| [\nabla F_i(w_i)]_j - [\nabla F(w)]_j || = (\sigma_{global})_j^2$
 for all $w_i\in R^d, j\in [d]$ and $i \in [m]$.
\end{assumption}
Assumptions~\ref{as1} and~\ref{as2} are widely used in nonconvex optimization literature~\cite{reddi2018adaptive,ward2018adagrad,li2019convergence}.
Assumption~\ref{as3} also has been used throughout various works on federated optimizations.
We adopt Assumption~\ref{as3} from~\cite{reddi2020adaptive}. 
Assumption~\ref{as3} bounds the variance among the client objective functions and the variance between the client objective function and the overall objective function.
In addition, with a slight abuse of notation, we use $\sigma_{l}^2$ and $\sigma_{g}^2$ to denote $\sum_{j=1}^{d}(\sigma_{local})_{i,j}^2$ and $\sum_{j=1}^{d}(\sigma_{global})_{i,j}^2$ respectively.

We first present a lemma that bounds the divergence of the client model $w_{i,k}^t$ to the overall model $w_t$ among all workers, which is essential in our theory.
\begin{lemma}(Bounded Model Divergence)\label{lemma:1}
Let Assumptions~\ref{as1},~\ref{as2} and~\ref{as3} hold, and any step size $\eta_l \leq \frac{1}{\sqrt{5}KL}$, we can bound the divergence of the client model $w_{i,k}^t$ to the overall model $w_t$ among all workers for any $k\in \{1, \cdots K-1\}$ as

CASE I: PreFed (Algorithm~\ref{alg:prefed})
\begin{equation*}
\small
\begin{split}
\frac{5}{KL^2\tau^2}\sum_{j=1}^{d} [(\sigma_{\text{local}})_{j}^2 + 5K(\sigma_{\text{global}})_{j}^2] + \frac{25}{L^2\tau^2}\mathbb E ||\nabla F(w_t)||^2
\end{split}
\end{equation*}

CASE II: PreFedOpt (Algorithm~\ref{alg:prefedopt})
\begin{equation*}
\small
\begin{split}
\frac{5}{L^2K}\sum_{j=1}^{d} [(\sigma_{\text{local}})_{j}^2 +5K(\sigma_{\text{global}})_{j}^2] + \frac{25}{L^2}\mathbb E ||\nabla F(w_t)||^2`
\end{split}
\end{equation*}
\end{lemma}
The proof is deferred to appendix B.
The above lemma shows that the divergence of client and overall model is mainly controlled by $\sigma_{\text{local}}$ and $\sigma_{\text{global}}$. Compared to PreFedOpt, PreFed uses local adaptivity for client model update, and the model divergence is scaled to $\frac{1}{\tau^2}$. Next, we provide the convergence analysis of PreFed.

\begin{theorem}\label{th:1}(Convergence of PreFed)
Let assume Assumptions~\ref{as1},~\ref{as2} and~\ref{as3} hold, and let $L, G, d,\sigma_{\text{local}}, \sigma_{\text{global}}$ be as defined therein. Let $\sigma^2 = \sum_{j=1}^{d} [(\sigma_{\text{local}})_{j}^2 +(5K+\frac{3}{2m})(\sigma_{\text{global}})_{j}^2]$. Assume the client models perform $T$ global synchronizations for every $K$ local updates. For any step size $\eta$, then the Algorithm~\ref{alg:prefed} for PreFed satisfy
\begin{equation}
\begin{split}
    &\min_{0\leq t\leq T-1} \mathbb E||\nabla F(x_t)||^2\\
    &\leq \mathcal O\bigg(\frac{1}{T} \Big[C_1 + G^2\big(\frac{L\eta + 3}{75}\big)m\Big]\bigg)
\end{split}
\end{equation}
where
\begin{equation}
    C_1 = \frac{2m}{75\eta}(F(w_0) - F(w_{T-1})) + \frac{L}{5K} \sigma^2
\end{equation}
\end{theorem}
To obtain an explicit dependence on $T$ and $K$, we simplify the above result for a specific choice of step size $\eta$.
\begin{corollary}
Suppose local step size $\eta_l$ in Lemma~\ref{lemma:1} is satisfied. Suppose $\eta$ satisfies $\eta= \Theta(2\sqrt{m}/75\sqrt{K})$. The for sufficiently large $T$, the Algorithm~\ref{alg:prefed} for for PreFed satisfy
\begin{equation}
\small
\begin{split}
    &\min_{0\leq t\leq T-1} \mathbb E||\nabla F(x_t)||^2\\
    &=\mathcal O\bigg( \frac{F(w_0) - F(w_{T-1})}{T\sqrt{Km}} + \frac{L\sigma^2}{5TK} + \frac{mG^2}{25T} + \frac{2Lm^{3/2}G^2}{75^2\sqrt{K}T}\bigg)
\end{split}
\end{equation}
\end{corollary}
The convergence analysis for PreFedOpt is different from PreFed because we only perform server-side adaptivity. We follow~\cite{reddi2020adaptive} work with the general non-convex setting of our interest and provide PreFedOpt convergence.
\begin{theorem}\label{th:2}(Convergence of PreFedOpt)
Let assume Assumptions~\ref{as1},~\ref{as2} and~\ref{as3} hold, and let $L, G, d,\sigma_{\text{local}}, \sigma_{\text{global}}$ be as defined therein.  Let $\sigma^2 = \sum_{j=1}^{d} [(\sigma_{\text{local}})_{j}^2 +5K(\sigma_{\text{global}})_{j}^2]$. Assume the client models perform $T$ global synchronizations for every $K$ local updates.The Algorithm~\ref{alg:prefedopt} for for PreFedOpt satisfy
\begin{equation}
\begin{split}
    &\min_{0\leq t\leq T-1} \mathbb E||\nabla F(x_t)||^2\\
    &\leq \mathcal O \Big(\big(\frac{G}{\sqrt{T}} + \frac{\tau}{\eta KT}\big)\big[C_1 + C_2 \times\min\{C_3, C_4\}\big]\Big)
\end{split}
\end{equation}
where
\begin{equation*}
\small
\begin{split}
    C_1 &= \frac{F(w_0) - F(w^*)}{\eta} + \frac{5\eta}{2}\sigma^2\\
    C_2 &= \frac{\eta KG^2}{\tau} + \frac{\eta L}{2}\\
    C_3 &= d + \sum_{j=1}^d\log(1 + \frac{\eta K^2 G^2 T}{\tau^2})\\
    C_4 &= \frac{16\eta K T}{m\tau} \sigma_{local}^2  + (16\eta^4K^3L^2T + 4\eta^2K^2T) \sigma^2
\end{split}
\end{equation*}
\end{theorem}
Similar to the PreFed case, we restate the above result for a specific choice of step size $\eta$ in order to highlight the dependence of $K$ and $T$.
\begin{corollary}
Suppose local step size $\eta_l$ in Lemma~\ref{lemma:1} is satisfied. Suppose $\eta$ satisfies $\eta= \Theta(\sqrt{GKm})$. Then for sufficiently large $T$, the Algorithm~\ref{alg:prefedopt} for PreFedOpt satisfy
\begin{equation}
\small
\begin{split}
    &\min_{0\leq t\leq T-1} \mathbb E||\nabla F(x_t)||^2\\
    &=\mathcal O\bigg(\frac{F(w_0) - F(w_{T-1})}{\sqrt{GKmT}} + \frac{5\sqrt{GKm}\sigma^2}{2T} +\frac{\sqrt{GKm}L}{2T} \times C_5 \bigg)
\end{split}
\end{equation}
where $C_5 = \min\{C_3, C_4\}$, $C_3$ and $C_4$ are defined in the Theorem~\ref{th:2} respectively.
\end{corollary}
\section{Experiments}
\begin{figure*}[!ht]
\centering
\includegraphics[width=1.\textwidth]{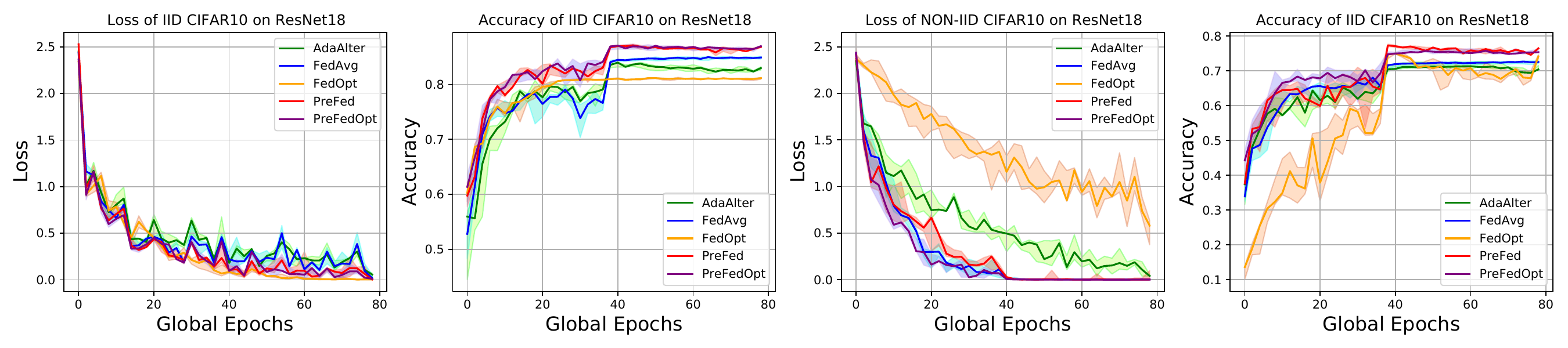}
\caption{Loss and validation accuracy of adaptive federated methods in ReNet18 on i.i.d. and non-i.i.d. CIFAR10 (best see in color).}
\label{cifar10}
\end{figure*}
\begin{figure*}
\centering
\includegraphics[width=1.\textwidth]{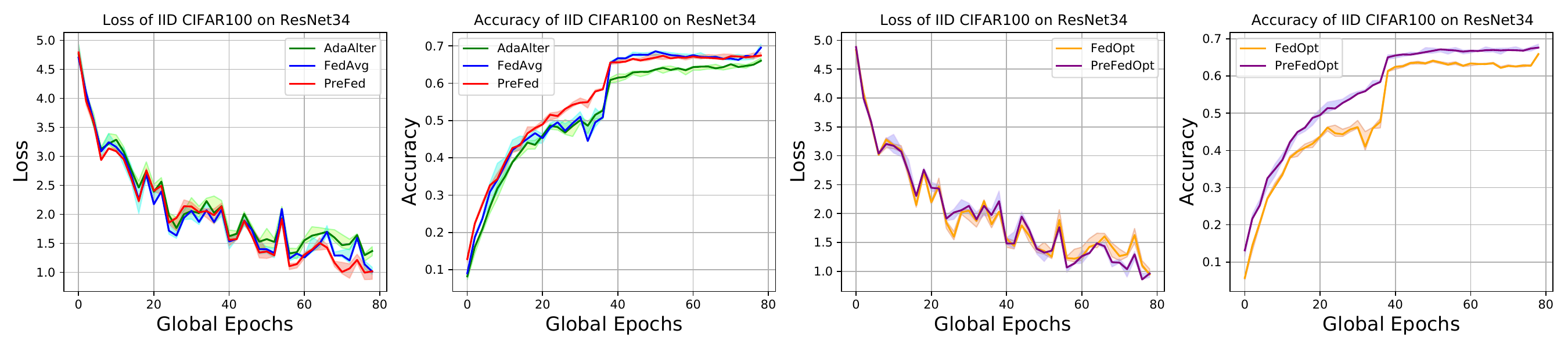}
\caption{Loss and validation accuracy of adaptive federated methods in ReNet34 on i.i.d. CIFAR100 (best see in color).}
\label{cifar100}
\end{figure*}
In this section, we present empirical evaluations of PreFed and PreFedOpt on different FL settings. We compare the performances of our methods to state-of-the-art approaches such as AdaAlter~\cite{xie2019local} and FedAdam~\cite{reddi2020adaptive}. We highlight the benefits of introducing both client-side and server-side preconditioning to federated optimization.

\textbf{Datasets and models}.
We use two popular image classification datasets: CIFAR-10 and CIFAR-100~\cite{cifar100}. 
The models we use for training are ResNet18 and ResNet34~\cite{he2016deep}, respectively.
We train models under two settings: i.i.d. and non-i.i.d.
Clients are sampled uniformly at random, without replacement in a given round, but with replacement across rounds. 
We implement all algorithms in the latest Pytorch 1.9 and provide an open-source implementation of all optimizers and benchmark tasks.

\textbf{Algorithms and hyperparameters}
We train all models in $E= 80$ global epochs. 
For each global epoch, we perform the global model synchronization for every $K=500$ iterations.
For FedAvg, we use SGD with momentum as a local optimizer with step size $\eta_l=0.05$ and momentum $\beta=0.9$.
For AdaAlter and FedAdam, we set $\eta_l=0.001, \beta_1=0.9, \beta_2=0.99$, and $\tau=0.001$ which are the same as standard Adam settings.
For PreFed and PreFedOpt, we set $\beta_2=0.9$.
The server-side step size a in FedAdam and PreFedOpt, we use $\eta_g=0.05$.
We decay the local step size by 0.1 for every 20 global epochs, and we decay server-side step size by 0.1 for every 40 global epochs.

\subsection{Results}
The two sub-figures on the left of Figure~\ref{cifar10} show the CIFAR10 on ResNet18 with i.i.d. settings. Both proposed algorithms perform better and achieve 88\% and 87\% accuracy on the validation dataset, respectively.
The two sub-figures (on the right-side) display interesting results of different methods on the non-i.i.d. dataset.  
Our proposed methods outperform AdaAlter and FedAdam, both methods achieving 78\% accuracy.
We notice that the standard FedAvg has a large variance during the training (see the light blue region). 
This is mainly due to the data heterogeneity. On the other hand, adaptive methods reduce the noise of stochastic gradients and are more stable.

Experiments on CIFAR100 (Figure~\ref{cifar100}) show our algorithms continuously improve the model performance. 
%Under the i.i.d. settings, proposed methods outperform the AdaAlter and FedOpt by $\sim 2\%$ to $\sim 4\%$.
Training with non-i.i.d settings of CIFAR100 is challenging because of the client models' variance. 
When the number of clients increases, each client contains only a small part of the dataset, the gradient methods such as SGD causing significant model divergence. 
We will leave more experiment results in Appendix B.

\section{Conclusion}
In this paper, we present the preconditioned federated algorithms, PreFed and PreFedOpt, to improve the learning efficiency for FL optimization. We theoretically analyze their convergence behavior. Our experiments show proposed algorithms achieve state-of-the-art performances.
\bibliography{aistats.bib}
\end{document}